%% file: main.tex
\documentclass[letterpaper, 10 pt, conference]{ieeeconf}  % Comment this line out
                                                         
\IEEEoverridecommandlockouts                              % This command is only
\usepackage{amsmath}
\usepackage{url}
\usepackage{graphicx}
\usepackage{booktabs} % For formal tables
\usepackage{graphicx}
\usepackage{subfig}
\usepackage{color-edits}
\usepackage{balance}
\usepackage{color, colortbl}
\usepackage{csvsimple}
\definecolor{LightCyan}{rgb}{0.88,1,1}
\title{\LARGE \bf
Predicting  Cyber Events by Leveraging Hacker Sentiment}

\author{Ashok Deb, Kristina Lerman, Emilio Ferrara
\\
USC Information Sciences Institute, Marina del Rey, CA
\\
{Email: \{{ashok, lerman, ferrarae\}}@isi.edu}
}

\begin{document}
\maketitle
\thispagestyle{empty}
\pagestyle{empty}

%%%%%%%%%%%%%%%%%%%%%%%%%%%%%%%%%%%%%%%%%%%%%%%%%%%%%%%%%%%%%%%%%%%%%%%%%%%%%%%%
\begin{abstract}

Recent high-profile cyber attacks exemplify why organizations need better cyber defenses. Cyber threats are hard to accurately predict because attackers usually try to mask their traces. However, they often discuss exploits and techniques on hacking forums. The community behavior of the hackers may provide insights into groups' collective malicious activity. We propose a novel approach to predict cyber events using sentiment analysis. We test our approach using cyber attack data from 2 major business organizations. We consider 3 types of events: malicious software installation, malicious destination visits, and malicious emails that surpassed the target organizations' defenses. We construct predictive signals by applying sentiment analysis on hacker forum posts to better understand hacker behavior. We analyze over 400K posts generated between January 2016 and January 2018 on over 100 hacking forums both on surface and Dark Web. We find that some forums have significantly more predictive power than others. Sentiment-based models that leverage specific forums can outperform state-of-the-art deep learning and time-series models on forecasting cyber attacks weeks ahead of the events.  
\end{abstract}

%%%%%%%%%%%%%%%%%%%%%%%%%%%%%%%%%%%%%%%%%%%%%%%%%%%%%%%%%%%%%%%%%%%%%%%%%%%%%%%%
\input{src/intro}
\input{src/data}
\input{src/sentiment}
\input{src/meth}
\input{src/res}
\input{src/rw}

\input{src/concl}

\section*{ACKNOWLEDGMENT}
{\footnotesize
This work has been partly funded by the  Intelligence Advanced Research Projects Activity (IARPA). The views and conclusions contained herein are those of the authors and should not be interpreted as necessarily representing the official policies, either expressed or implied, of IARPA, or the U.S. Government. The U.S. Government had no role in study design, data collection and analysis, decision to publish, or preparation of the manuscript. The U.S. Government is authorized to reproduce and distribute reprints for governmental purposes notwithstanding any copyright annotation therein.
}

\addtolength{\textheight}{-5cm}   % This command serves to balance the column lengths

%%%%%%%%%%%%%%%%%%%%%%%%%%%%%%%%%%%%%%%%%%%%%%%%%%%%%%%%%%%%%%%%%%%%%%%%%%%%%%%%

%%%%%%%%%%%%%%%%%%%%%%%%%%%%%%%%%%%%%%%%%%%%%%%%%%%%%%%%%%%%%%%%%%%%%%%%%%%%%%%%

%%%%%%%%%%%%%%%%%%%%%%%%%%%%%%%%%%%%%%%%%%%%%%%%%%%%%%%%%%%%%%%%%%%%%%%%%%%%%%%%

%%%%%%%%%%%%%%%%%%%%%%%%%%%%%%%%%%%%%%%%%%%%%%%%%%%%%%%%%%%%%%%%%%%%%%%%%%%%%%%%

\bibliographystyle{ieeetr}
\bibliography{acmart}

\end{document}

%% file: src/intro.tex
\section{Introduction}
Recent high-profile cyber attacks---the massive denial of service attack using Mirai botnet, infections of computers word-wide with WannaCry and Petya ransomware, the Equifax data breach---highlight the need for organizations to develop cyber crime defenses. Cyber threats are hard to identify and predict because the hackers that conduct these attacks often obfuscate their activity and intentions. However, they may still use publicly accessible forums discuss vulnerabilities and share tradecraft about how to exploit them. The behavior of the hacker community, as expressed in such venues, may provide insights into group's malicious intent. It has been shown that computational models based on various behavior learning theories can help in cyber security situational awareness \cite{dutt2013cyber}. While cyber situation awareness \cite{jajodia2009cyber, franke2014cyber} is critical for defending networks, it is focused on detecting cyber events. %the present state. 
In this paper, we describe a computational method that analyzes discussions on hacker forums  to predict cyber attacks.

Opinion mining or sentiment analysis can be linked all the way back to Freud's 1901 paper on how slips of the tongue can reveal a person's hidden intentions \cite{freud19011960}. While sentiment analysis was originally developed in the field of linguistics and psychology, it has recently been applied to a number of other fields with the first seminal work in the computational sciences being Pang 2002 \cite{pang2002thumbs}. Historically, the context it has been applied to are social networks, comments (such as on news sites) and reviews (either for products or movies). In this work, we apply sentiment analysis to posts on Dark Web forums with the purpose of forecasting cyber attacks. The Dark Web consists of websites that are not indexed nor searchable by standard search engine and can only be accessed using a special browser service. 

We further explore the link between community behavior and malicious activity. The connection between security and human behavior has been studied in designing new technology \cite{pfleeger2012leveraging}, here we look to reverse engineer by mapping the malicious events to hacker behavior. Social media has been shown to be a source of useful data on human behavior and used to predict real world events \cite{agarwal2015applying,asur2010predicting,kalampokis2013understanding}. Here, we inspect the ability of hacker forums to predict cyber events.
We consider each individual forum, applying sentiment analysis to each post in the forum. After computing a daily average per forum and a 7 day running average sentiment signal per forum, we test these signals against ground truth data. We determine some forums have significantly more predictive power and these isolated forums can beat the evaluation models in 36\% of the months under study using precision and recall of predictions within a 39-hour window of the event.

% \subsection*{Contributions of this work}

% \begin{itemize}
% \item ...
% \end{itemize}

%% file: src/data.tex
\section{Data}
\subsection{Hacker Forum Texts}
We look at hacking forums from both the Surface Web and the Dark Web from 1 January 2016 to 31 January 2018. The Dark Web refers to sites accessible through The Onion Router private network platform \cite{dingledine2004tor}. The Surface Web refers to the World Wide Web accessible through standard browsers. In this paper, we focus only on English posts from 113 forums which were identified based on cyber security keywords consisting of 432,060 posts. The text from these forums were accessed using the methods proposed in \cite{robertson2017darkweb,nunes2016darknet}.

%\textbf{chart of darkweb post by forum}
\subsection{Ground truth data}
% * <adeb@usc.edu> 2018-04-08T19:03:53.548Z:
% 
% Can we cite the IARPA CAUSE Ground Truth Handbook?
% 
% ^ <adeb@usc.edu> 2018-04-08T19:04:25.078Z.

%This is important because there is a strong need to protect the DIB as outlined as part of the US National Strategy \cite{house2003national}.
We use ground truth data of cyber attacks from 2 major organizations in the Defense Industrial Base (DIB) industry.  Henceforth, we will refer to them as Organization A and Organization B for anonymity. The ground truth comprises 3 event types:
\begin{itemize}
\item \textit{endpoint-malware:} a malicious software installation such as ransomware, spyware and adware is discovered on a company endpoint device. 
\item \textit{malicious-destination} a visit by a user to a URL or IP address that is malicious in nature or a compromised website. 
\item \textit{malicious-email} receipt of an email that contains a malicious email attachment and/or a link to a known malicious destination. 
\end{itemize}

%% file: src/sentiment.tex
\section{Sentiment Analysis}
The first effective use of sentiment analysis in a predictive sense was by Pang et. al.  \cite{pang2002thumbs} in assessing movie reviews. Since then, sentiment analysis has expanded to other fields. Sentiment analysis can be done with or without supervision (label training data). Supervised methods can be adapted to create trained models for specific purposes and contexts. The drawback is that labeled data may be highly costly and often researchers end up using AMT - Amazon Mechanical Turk. The alternative is to use lexical-based methods that do not rely on labeled data; however, it is hard to create a unique lexical-based dictionary to be used for all different contexts. Deep learning methods allow for additional functions like taking into account order of words in a sentence like the Stanford Recursive Deep Model. Methods can either be 2 way (positive or negative) or 3 way (positive, neutral, negative). Furthermore, dictionary based sentiment algorithms are either polarity-based where sentiment is based only of the frequency of positive or negative words whereas valence-based methods factor the intensity of the words into polarity. 
There are a number of issues with sentiment analysis which include: word pairs, word tuples, emoticons, slang, sarcasm, irony, questions, URLs, code, domain specific use of words (shoot an email, dead link), and inversions (small is good for portable electronics) which are difficult for computerized text analysis to handle. 

Studies have found that a method’s prediction performance varies considerably from one dataset to another. VADER works well for some tweets, but not for others, depending on the context. SentiStrength has good Macro F1 values, but has low coverage because it tends to classify a high number of instances as neutral. 

The choice of a sentiment analysis is highly dependent on the data and application, therefore you need to take into account prediction performance and coverage. There is no single method that always achieves a consistent rank position for different datasets. Therefore, in this paper we test multiple methods for sentiment analysis. Most languages themselves are biased positive and if a lexicon is built on data, the positive bias that data can lead to a bias in the lexicon. This is why most methods are better at classifying positive than neutral or negative methods meaning that they are biased, neutral are the hardest to detect \cite{sentibench}.

\subsection{Vader}
VADER: Valence Aware Dictionary for sEntiment Reasoning \cite{vader} is a rule-based sentiment model that has both a dictionary and associated intensity measures. It's dictionary has been tuned for microblog-like contexts and they incorporate 5 generalizable rules that goes beyond pure dictionary lookups:
\begin{enumerate}
\item Increase intensity due to exclamation point

\item Increase intensity due to all caps in the presence of other non-all cap words
\item  Increase intensity with degree modifiers i.e. extremely
\item  Negate sentiment with contrastive conjunction i.e. but
\item  Examine the preceding tri-gram to identify cases where negation flips the polarity of the text. 
\end{enumerate}

Therefore, VADER not only captures positive or negative, but also how positive and how negative beyond simple words counts. It is made further robust by the additional rules. It's "gold standard" lexicon was developed manually and with Amazon Mechanical Turk. Vader scores range from 0.0 to 1.0.

\subsection{LIWC}
Linguistic Inquiry and Word Count (LIWC) \cite{pennebaker2001linguistic} was a pioneer in the  computerized text analysis field with the first major iteration in 2007, we used the updated version LIWC 2015. It has two components: the processing component and the dictionaries. The heart of LIWC are the dictionaries that contain the lookup words in psychometric categories which is able to resolve content words from style words. LIWC counts the inputted words in psychologically meaningful categories which produces close to 100 dimensions for any given text being analyzed. For the purposes of this research, we are only focused on Tone which bests maps to sentiment as we have defined it. The Tone scores range from 0 to 100. LIWC also ignores context, irony, sarcasm, and idioms.

\subsection{SentiStrength}
SentiStrength \cite{senti} is another lexicon-based sentiment classifier which leverages dictionaries and non-lexical linguistics information to detect sentiment. SentiStrength focuses on the strength of the sentiment and uses weights for the words in its dictionaries. Additionally, positive sentiment strength and negative sentiment strength is scored separately. Each is scored from 1 to 5, with 5 being the greatest strength. For our purposes, we seek overall sentiment so we subtract the negative sentiment from the positive sentiment so that strongly positive (5,1) becomes 4, neutral (1,1) becomes 0 and strongly negative (1,5) becomes -4. Therefore, SentiStrength scorese range from -4 to 4. SentiStrength is designed to do better with social media; however, it can't exploit indirect indicators of sentiment. It is also weaker for positive sentiment in news-related discussions. 

%% file: src/meth.tex
\section{Methodology}
In this section we document the methodology used and process workflow from the data processing to signal generation through warning generation and signal testing. Three cases studies are used to illustrate the process via example and Figure X provides a visual reference. 

\subsection{Processing the Data}
Working with researchers at Arizona State University, we were able to develop a database of posts from forums on both the Dark Web and Surface Web which discuss computer security and network vulnerability topics. To protect the future utility of these sources, each forum has been coded with a number (forumid) from 1 to 350. The data consist of the forumid, date the post was made, and the text of the post. The data in this study was from 1 January 2016 to 31 January 2018. The data was collected by ASU and we used an API to pull and store the data in a local server and access it via Apache Lucene's Elastic Search engine.

\subsection{Evaluating Sentiment Analysis}
After a review of the sentiment analysis methods in SentiBench \cite{sentibench}, we decided to use Vader\cite{vader}, SentiStrenght\cite{senti} and LIWC15\cite{pennebaker2001linguistic}. For social networks, VADER and LIWC15 were found to be the best method for 3-class classification and SentiStrength was the winner for 2-class classification. \cite{sentibench} These three methods were used because they  Vader has a Python module, SentiStrenght has a Java implementation and LIWC15 is a stand-alone program. 
\subsection{Computing Daily Averages}
A sentiment score for each forum post was computed using the three sentiment methods outlined above. Since there can be multiple posts on a forum for a day, we characterization the overall sentiment of the day with a daily average. There can be a wide range of sentiment scores for any given day, especially if there are a lot of posts from on a popular forum. In order to understand the trend of sentiment over time, we compute running averages. 

\subsection{Computing Running Averages}
\begin{figure}[!t]
  \includegraphics[width=\linewidth]{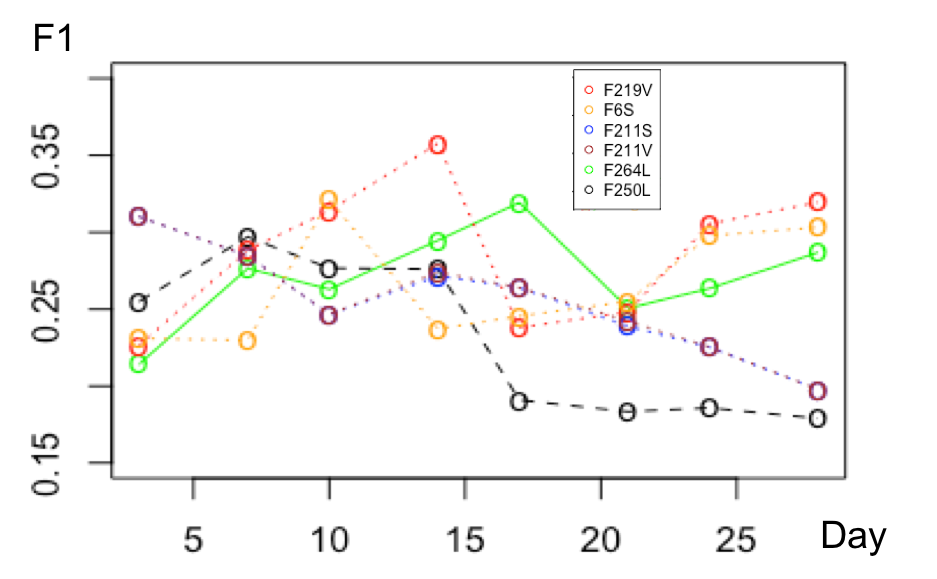}
  \caption{Average F1 Scores by Signal using Different Running Averages}
  \label{fig:avgs}
\end{figure}

A running daily average was computed in order to assess the trend of sentiment over time. The more days in the running average, the smoother the curve and the harder to detect a change. Whereas no using a running average or making it only 1 or 2 days would have many jump discontinuities and swings. We looked at adjusting the running average from 1 to 30 days and settled on 7 days primarily because that was our original prediction window. Figure \ref{fig:avgs} shows the average F1 score various signals computed with running averages of 3, 7, 10 and 14 days.  

% Figure \ref{fig:avgs} illustrates a same table visually, based on our analysis, we decided to use 7 days for the running average.  

%below table provides the same information in the graph
% \begin{table}[!t]
% \centering
% \caption{Average F1 for Signals Using Different Running Averages   \label{tab:avgs}} 
% \begin{tabular}{lcccc}%
% \hline%
% \textbf{Signal}&\textbf{3\_day}&\textbf{7\_day}&\textbf{10\_day}&\textbf{14\_day}\\%
% \hline%
% F6-senti	&	0.2496	&	0.2365	&	0.2620	&	0.2520	\\%
% F8-senti	&	0.0649	&	0.0651	&	0.0681	&	0.0793	\\%
% F8-vader	&	0.0649	&	0.0651	&	0.0684	&	0.0793	\\%
% F211-senti	&	0.2399	&	0.2436	&	0.2297	&	0.2528	\\%
% F211-vader	&	0.2067	&	0.1994	&	0.2008	&	0.2121	\\%
% F219-vader	&	0.2555	&	0.2918	&	0.3042	&	0.3283	\\%
% F250-LIWC	&	0.1902	&	0.1896	&	0.1937	&	0.1947	\\%
% F264-LIWC	&	0.2008	&	0.1927	&	0.1944	&	0.1988	\\%
% \hline%
% \end{tabular}%
% \end{table}

\subsection{Standardizing the Score}
To make the 3 sentiment scores more comparable, their scores were standardized. As previously mentioned, VADER generates sentiment scores on a scale of 0 to 1, SentiStrength goes from -4 to 4, and LIWC goes from 0 to 100 for Tone. While standardizing the scores do not affect the correlation any forum would have with the ground truth from our target organizations, it will be necessary when we potentially combine signals from various forums and sentiment methods to find more powerful predictors. 

\subsection{Compute Correlations to Find Potential Signals}
As previously mention, we have ground truth events from 2 defense industrial base organizations of 3 different cyber event types. The event types are endpoint-malware, malicious-destination and malicious-email. Correlations were computed between all forum-sentiments against all event types from both organizations. Additionally, since we are looking for predictive signals, we computed correlations with a negative lag from 0 to 30 days with a lag of -30 meaning offset the sentiment signal 30 days before the organization's event occurrence. A number of signals stood out as being more correlated than others against certain event types as seen in Figure \ref{tab:corr}. This shows the LIWC sentiment on Forum 84 against Organization B's endpoint-malware events. The fact that multiple, consecutive lags have low p-values gives some indication that this might be a useful signal.

\subsection{Forecasting Models}
We also apply widely-used ARIMA model for forecasting events.
ARIMA stands for autoregressive integrated moving average. The key idea is that
the number of current events ($y_t$) depends on the past counts and forecast
errors. Formally, ARIMA($p$,$d$,$q$) defines an autoregressive model with $p$
autoregressive lags, $d$ difference operations, and $q$ moving average lags
(see~\cite{shumway2010time}). Given the observed series of events
$\mathcal{Y}=(y_1,y_2,\ldots,y_T)$, ARIMA($p$,$d$,$q$) applies $d$ ($\ge 0$)
difference operations to transform $\mathcal{Y}$ to a stationary series
$\mathcal{Y}^{\prime}$. Then the predicted value $y^\prime_t$ at time point
$t$ can be expressed in terms of past observed values and forecasting errors
which is as follows: 
\begin{equation}
y^\prime_t = \mu_y + + \sum_{i=1}^p \alpha_i y^\prime_{t-i} + \sum_{j=1}^q \beta_j e_{t-j} + e_t\label{eq:arima}
\end{equation}

Here $\mu_y$ is a constant, $\alpha_i$ is the autoregressive (AR) coefficient at lag $i$, $\beta_j$ is the moving average (MA)
coefficient at lag $j$, $e_{t-j} = y^\prime_{t-j} - \hat{y}^\prime_{t-j}$ is the forecast error at lag $j$, and $e_t$ is assumed to be the white noise ($e_t\sim \mathcal{N}(0,\sigma^2)$).  The
AR model is essentially an ARIMA model without moving average terms. 

We use maximum likelihood estimation for learning the parameters; more
specifically, parameters are optimized with LBFGS
method~\cite{seabold2010statsmodels}. These models assume that $(p, d, q)$ are
known and the series is weakly stationary. To select the values for $(p, d, q)$
we employ grid search over the values of $(p, d, q)$ and select the one with
minimum AIC score.

\subsection{Testing Signals with ARIMAX}
Again, Table \ref{tab:corr} shows the signals that are better correlated with Organization B's ground truth events. The next step is to test these signals to see if they have any predictive power. To do this, the ARIMA model is used with the ground truth events to develop a baseline model from which to compare potential signals for the potential to have predictive power. Additionally, 4 other methods were used for comparison: Dark-Mentions, Deep-Exploit \cite{sapienza2017early}, ARIMAX with abuse.ch and a daywise-hourly-baserate model. Using ground truth events from both Organization A and Organization B, sentiment signals from the various forums, computed with the different methodologies were tested. Testing was done across the 3 event types for both Organizations with Precision, Recall and F1 computed to evaluate the signal. The timeseries of the sentiment for a given forum and sentiment method was used as the input to the timeseries forecasting model to predict future events. The model was trained on data from April 1, 2016 to May 31, 2017, in order to start generating warnings for the month of June 2017. After predictions were made for the month of June, they were scored against the actual ground truth and then the model was ran again to predict warnings for August 2017. This was done for all the way through January 2018.

\begin{table}[!t]
\centering
\caption{Best Signals for Organization B's Events
\label{tab:corr}} 
\begin{tabular}{lrcccr}
\hline%
\textbf{Forum\#}&\textbf{Sent}&\textbf{Lag}&\textbf{Correlation}&\textbf{p Value}&\textbf{Events}\\%
\hline%

84	&	LIWC	&	-11	&	0.2170	&	0.000055	&	EP-Mal	\\
84	&	LIWC	&	-12	&	0.2221	&	0.000037	&	EP-Mal	\\
84	&	LIWC	&	-14	&	0.2185	&	0.000052	&	EP-Mal	\\
219	&	Vader	&	-18	&	-0.2329	&	0.000079	&	EP-Mal	\\
264	&	LIWC	&	-10	&	0.2472	&	0.000040	&	EP-Mal	\\
264	&	LIWC	&	-12	&	0.2362	&	0.000095	&	EP-Mal	\\
264	&	LIWC	&	-15	&	0.2380	&	0.000091	&	EP-Mal	\\
159	&	Senti	&	-14	&	0.8498	&	0.000008	&	Mal-Email	\\
266	&	Senti	&	-14	&	-0.5517	&	0.000058	&	Mal-Email	\\
261	&	LIWC	&	-3	&	0.2173	&	0.000043	&	Mal-Dest	\\
266	&	Senti	&	-27	&	-0.6243	&	0.000080	&	Mal-Dest	\\

\hline%

\end{tabular}%
\end{table}

\subsection{Scoring}
To determine how well the signals under study performed, a matching algorithm was used to compare the date occurrence of the predicted events with the actual events that occurred. Using the matching algorithm, we could consistently score which predicted events should be mapped to actual events and which predicted events did not occur as well as which actual events were not predicted. There is a window around the actual events which varies based on the event type. Endpoint-maleware has to be within 0.875 days, malicious-destination within 1.625 days and malicious-email within 1.375 days.

\subsection{External Signals}
Currently, there are other external signals that the data provider Organizations are currently evaluating for predictive potential. Again, external signals are timeseries information derived from open sources that are not based on information system network data. The other external signals under evaluation are: 
\begin{itemize}
\item ARIMAX: is the same model outlined in \S 4.7; however, time series counts of malicious activity are acquired from \url{https://abuse.ch} and used in conjunction with historical data. 
\item Baseline: is the exact same model in \S 4.7 with no external signal and using only historical ground truth data to predict the future rate of attack.
\item Daywise-Baserate: is the same as the ARIMAX model mentioned above; however, the model takes day of the week into consideration assuming that the event rate for each day of the week is not the same. 
\item Deep-Exploit: is an ARIMA model that is based on the vulnerability analysis determined by \cite{tavabi2018dark}. This method referred to as DarkEmbed learns the embeddings of Dark Web posts and then uses a trained exploit classifier to predicted which vulnerabilities in Dark Web posts might be exploited. 
\item Dark-Mentions: Is an extension of \cite{almukaynizi2017proactive} which predicts if a disclosed vulnerability will be exploited based on a variety of data sources in addition to the Dark Web using methods still being developed. These predictions are used to construct a rule based forecasting method based on keyword mentions in Dark Web forums and marketplaces. 
\end{itemize}

%% file: src/res.tex
\section{Results}
After generating ARIMAX models with each potential signal, they were scored as mentioned above for each month from July 2017 to January 2018. The following tables show the results for the months under study. By month, you can see the number of actual ground truth events (Evt), the number of warnings generated by each signal (Warn), and the precision (P), recall (R) and F1 score for each. The table is sorted by largest F1 score for each month with only the top five signals listed. Signals generated by sentiment analysis that were part of the top five for each month are highlighted in light blue. 
\subsection{Organization A}
Table \ref{tab:output1} shows Organization A's endpoint-malware where sentiment signals dominated July, September and November and did reasonable well in the remaining months. Every month a sentiment signal beat at least on evaluation model. Malicious-Destination (Table \ref{tab:output2}) had periodic performance July, September, November and January but the case is not as strong as Endpoint-Malware. Lastly, Table \ref{tab:output3} shows Malicious-Email results which illustrate that sentiment signals did well in July to September with waning results for the later months. Upon further inspection this is believed to be due to some key forums going offline toward the end of the year.

\begin{table}[!t]
\centering
\caption{Results from Organization A's Endpoint-Malware
\label{tab:output1}} 
\begin{tabular}{llccccc}%
\hline%
\textbf{Month}&\textbf{Evt}&\textbf{Warn}&\textbf{Signal}&\textbf{P}&\textbf{R}&\textbf{F1}\\%
\hline%

\rowcolor{LightCyan}July	&	15	&	14	&	forum211-Senti		&	0.57	&	0.53	&	0.55	\\
\rowcolor{LightCyan}July	&	15	&	29	&	forum196-LIWC		&	0.41	&	0.80	&	0.55	\\
\rowcolor{LightCyan}July	&	15	&	27	&	forum89-Senti		&	0.41	&	0.73	&	0.52	\\
\rowcolor{LightCyan}July	&	15	&	12	&	forum111-LIWC		&	0.58	&	0.47	&	0.52	\\
July	&	15	&	9	&	baseline		&	0.67	&	0.40	&	0.50	\\
\hline
August	&	19	&	14	&	baseline		&	0.71	&	0.53	&	0.61	\\
\rowcolor{LightCyan}August	&	19	&	11	&	forum111-LIWC		&	0.82	&	0.47	&	0.60	\\
\rowcolor{LightCyan}August	&	19	&	35	&	forum8-Vader		&	0.46	&	0.84	&	0.59	\\
August	&	19	&	8	&	daywise-baserate		&	1.00	&	0.42	&	0.59	\\
\rowcolor{LightCyan}August	&	19	&	23	&	forum230-Senti		&	0.52	&	0.63	&	0.57	\\
\hline
\rowcolor{LightCyan}September	&	18	&	16	&	forum111LIWC		&	0.69	&	0.61	&	0.65	\\
\rowcolor{LightCyan}September	&	18	&	32	&	forum250LIWC		&	0.50	&	0.89	&	0.64	\\
\rowcolor{LightCyan}September	&	18	&	35	&	forum211vader		&	0.46	&	0.89	&	0.60	\\
\rowcolor{LightCyan}September	&	18	&	41	&	forum147LIWC		&	0.41	&	0.94	&	0.58	\\
\rowcolor{LightCyan}September	&	18	&	41	&	forum194LIWC		&	0.41	&	0.94	&	0.58	\\
\hline
October	&	6	&	14	&	daywise-baserate		&	0.29	&	0.67	&	0.40	\\
October	&	6	&	35	&	baseline		&	0.17	&	1.00	&	0.29	\\
\rowcolor{LightCyan}October	&	6	&	29	&	forum8vader		&	0.17	&	0.83	&	0.29	\\
\rowcolor{LightCyan}October	&	6	&	37	&	forum111LIWC		&	0.16	&	1.00	&	0.28	\\
\rowcolor{LightCyan}October	&	6	&	43	&	forum211vader		&	0.14	&	1.00	&	0.24	\\
\hline
\rowcolor{LightCyan}November	&	27	&	38	&	forum6senti		&	0.63	&	0.89	&	0.74	\\
\rowcolor{LightCyan}November	&	27	&	42	&	forum147LIWC		&	0.60	&	0.93	&	0.72	\\
\rowcolor{LightCyan}November	&	27	&	40	&	forum111LIWC		&	0.60	&	0.89	&	0.72	\\
\rowcolor{LightCyan}November	&	27	&	41	&	forum211senti		&	0.59	&	0.89	&	0.71	\\
\rowcolor{LightCyan}November	&	27	&	43	&	forum121LIWC		&	0.56	&	0.89	&	0.69	\\
\hline
December	&	13	&	18	&	arimax		&	0.33	&	0.46	&	0.39	\\
December	&	13	&	16	&	dark-mentions		&	0.31	&	0.38	&	0.34	\\
\rowcolor{LightCyan}December	&	13	&	80	&	forum121LIWC		&	0.16	&	1.00	&	0.28	\\
\rowcolor{LightCyan}December	&	13	&	73	&	forum194LIWC		&	0.16	&	0.92	&	0.28	\\
December	&	13	&	10	&	deep-exploit		&	0.30	&	0.23	&	0.26	\\
\hline
January	&	1	&	15	&	dark-mentions		&	0.07	&	1.00	&	0.13	\\
\rowcolor{LightCyan}January	&	1	&	37	&	forum6senti		&	0.03	&	1.00	&	0.05	\\
\rowcolor{LightCyan}January	&	1	&	61	&	forum147LIWC		&	0.02	&	1.00	&	0.03	\\
January	&	1	&	64	&	baseline		&	0.02	&	1.00	&	0.03	\\
January	&	1	&	19	&	arimax		&	0.00	&	0.00	&	0.00	\\

\hline%
\end{tabular}%
\end{table}

\begin{table}[!t]
\centering
\caption{Results from Organization A's Malicious-Destination
\label{tab:output2}} 
\begin{tabular}{llllccc}%
\hline%
\textbf{Month}&\textbf{Evt}&\textbf{Warn}&\textbf{Signal}&\textbf{P}&\textbf{R}&\textbf{F1}\\%
\hline%

July	&	4	&	5	&	baseline		&	0.40	&	0.50	&	0.44	\\
July	&	4	&	3	&	daywise-baserate		&	0.33	&	0.25	&	0.29	\\
July	&	4	&	17	&	dark-mentions		&	0.12	&	0.50	&	0.19	\\
\rowcolor{LightCyan}July	&	4	&	42	&	forum266-LIWC		&	0.05	&	0.50	&	0.09	\\
July	&	4	&	0	&	arimax		&	0.00	&	0.00	&	0.00	\\
\hline 
August	&	10	&	6	&	baseline		&	1.00	&	0.60	&	0.75	\\
August	&	10	&	10	&	daywise-baserate		&	0.60	&	0.60	&	0.60	\\
August	&	10	&	8	&	dark-mentions		&	0.50	&	0.40	&	0.44	\\
August	&	10	&	0	&	arimax		&	0.00	&	0.00	&	0.00	\\
August	&	10	&	0	&	deep-exploit		&	0.00	&	0.00	&	0.00	\\
\hline
\rowcolor{LightCyan}September	&	4	&	15	&	forum194LIWC		&	0.20	&	0.75	&	0.32	\\
\rowcolor{LightCyan}September	&	4	&	15	&	forum210LIWC		&	0.20	&	0.75	&	0.32	\\
\rowcolor{LightCyan}September	&	4	&	15	&	forum264LIWC		&	0.20	&	0.75	&	0.32	\\
\rowcolor{LightCyan}September	&	4	&	15	&	forum6senti		&	0.20	&	0.75	&	0.32	\\
\rowcolor{LightCyan}September	&	4	&	15	&	forum194LIWC		&	0.20	&	0.75	&	0.32	\\
\hline
October	&	2	&	0	&	arimax		&	0.00	&	0.00	&	0.00	\\
October	&	2	&	0	&	dark-mentions		&	0.00	&	0.00	&	0.00	\\
October	&	2	&	5	&	daywise-baserate		&	0.00	&	0.00	&	0.00	\\
October	&	2	&	0	&	deep-exploit		&	0.00	&	0.00	&	0.00	\\
\hline
November	&	1	&	5	&	daywise-baserate		&	0.20	&	1.00	&	0.33	\\
\rowcolor{LightCyan}November	&	1	&	6	&	forum111LIWC		&	0.17	&	1.00	&	0.29	\\
\rowcolor{LightCyan}November	&	1	&	6	&	forum147LIWC		&	0.17	&	1.00	&	0.29	\\
\rowcolor{LightCyan}November	&	1	&	30	&	forum210senti		&	0.03	&	1.00	&	0.06	\\
November	&	1	&	0	&	arimax		&	0.00	&	0.00	&	0.00	\\
\hline
December	&	1	&	10	&	daywise-baserate		&	0.10	&	1.00	&	0.18	\\
December	&	1	&	11	&	dark-mentions		&	0.09	&	1.00	&	0.17	\\
December	&	1	&	0	&	arimax		&	0.00	&	0.00	&	0.00	\\
December	&	1	&	0	&	deep-exploit		&	0.00	&	0.00	&	0.00	\\
\hline
\rowcolor{LightCyan}January	&	2	&	24	&	forum111LIWC		&	0.08	&	1.00	&	0.15	\\
January	&	2	&	0	&	arimax		&	0.00	&	0.00	&	0.00	\\
January	&	2	&	10	&	dark-mentions		&	0.00	&	0.00	&	0.00	\\
January	&	2	&	9	&	daywise-baserate		&	0.00	&	0.00	&	0.00	\\
January	&	2	&	0	&	deep-exploit		&	0.00	&	0.00	&	0.00	\\

\hline%

\end{tabular}%
\end{table}

\begin{table}[!t]
\centering
\caption{Results from Organization A's Malicious-Email
\label{tab:output3}} 
%\begin{tabular}{llccccc}%
\begin{tabular}{llccccc}
\hline%
\textbf{Month}&\textbf{Evt}&\textbf{Warn}&\textbf{Signal}&\textbf{P}&\textbf{R}&\textbf{F1}\\%
\hline%
\rowcolor{LightCyan}July	&	26	&	21	&	forum210-LIWC	&	0.76	&	0.62	&	0.68	\\
\rowcolor{LightCyan}July	&	26	&	27	&	forum250-LIWC	&	0.67	&	0.69	&	0.68	\\
\rowcolor{LightCyan}July	&	26	&	19	&	forum147-LIWC	&	0.74	&	0.54	&	0.62	\\
\rowcolor{LightCyan}July	&	26	&	36	&	forum159-Senti	&	0.53	&	0.73	&	0.61	\\
\rowcolor{LightCyan}July	&	26	&	17	&	forum28-LIWC	&	0.76	&	0.50	&	0.60	\\
\hline
\rowcolor{LightCyan}August	&	11	&	17	&	forum179-Vader	&	0.59	&	0.91	&	0.71	\\
\rowcolor{LightCyan}August	&	11	&	15	&	forum250-LIWC	&	0.60	&	0.82	&	0.69	\\
August	&	11	&	7	&	daywise-baserate	&	0.86	&	0.55	&	0.67	\\
\rowcolor{LightCyan}August	&	11	&	18	&	forum210-Senti	&	0.50	&	0.82	&	0.62	\\
\rowcolor{LightCyan}August	&	11	&	25	&	forum159-Senti	&	0.44	&	1.00	&	0.61	\\
\hline
\rowcolor{LightCyan}September	&	15	&	36	&	forum264LIWC	&	0.36	&	0.87	&	0.51	\\
September	&	15	&	17	&	daywise-baserate	&	0.47	&	0.53	&	0.50	\\
\rowcolor{LightCyan}September	&	15	&	18	&	forum210senti	&	0.44	&	0.53	&	0.48	\\
\rowcolor{LightCyan}September	&	15	&	45	&	forum147LIWC	&	0.31	&	0.93	&	0.47	\\
\rowcolor{LightCyan}September	&	15	&	46	&	forum6senti	&	0.28	&	0.87	&	0.43	\\
\hline
October	&	11	&	14	&	daywise-baserate	&	0.50	&	0.64	&	0.56	\\
October	&	11	&	8	&	deep-exploit	&	0.50	&	0.36	&	0.42	\\
\rowcolor{LightCyan}October	&	11	&	42	&	forum264LIWC	&	0.17	&	0.64	&	0.26	\\
\rowcolor{LightCyan}October	&	11	&	51	&	forum194LIWC	&	0.16	&	0.73	&	0.26	\\
\rowcolor{LightCyan}October	&	11	&	102	&	forum8vader	&	0.11	&	1.00	&	0.19	\\
\hline
November	&	50	&	16	&	daywise-baserate	&	0.69	&	0.22	&	0.33	\\
November	&	50	&	4	&	deep-exploit	&	0.75	&	0.06	&	0.11	\\
November	&	50	&	0	&	arimax	&	0.00	&	0.00	&	0.00	\\
November	&	50	&	0	&	dark-mentions	&	0.00	&	0.00	&	0.00	\\
\hline
December	&	17	&	22	&	daywise-baserate	&	0.55	&	0.71	&	0.62	\\
December	&	17	&	10	&	deep-exploit	&	0.80	&	0.47	&	0.59	\\
December	&	17	&	5	&	dark-mentions	&	0.80	&	0.24	&	0.36	\\
December	&	17	&	0	&	arimax	&	0.00	&	0.00	&	0.00	\\
\hline
January	&	40	&	18	&	daywise-baserate	&	0.94	&	0.43	&	0.59	\\
January	&	40	&	8	&	deep-exploit	&	0.75	&	0.15	&	0.25	\\
January	&	40	&	6	&	dark-mentions	&	0.83	&	0.13	&	0.22	\\
January	&	40	&	0	&	arimax	&	0.00	&	0.00	&	0.00	\\

\hline%

\end{tabular}%
\end{table}

\subsection{Organization B}
Table \ref{tab:output4} shows that sentiment signals do best for July and October for Malicious-Destination. While baseline and daywise-baserate dominate the other months, sentiment signals perform better than the other evaluation models. 
Similar to Organization A, the Malicious-Destination for Organization B (Table \ref{tab:output5}) does the best early in July in August and moderately well in September to November  until degrading to below all evaluation models in December and January. This may be due the few number of events and perhaps sentiment signals do not perform the best under low frequency conditions. The performance for Malicious-Email (Table \ref{tab:output6}) is oddly cyclical; however, sentiment signals dominated December and beat at least one evaluation model for every month.

\begin{table}[!t]
\centering
\caption{Results from Organization B's Endpoint-Malware
\label{tab:output4}} 
%\begin{tabular}{llccccc}%
\begin{tabular}{llccccc}
\hline%
\textbf{Month}&\textbf{Evt}&\textbf{Warn}&\textbf{Signal}&\textbf{P}&\textbf{R}&\textbf{F1}\\%
\hline%

\rowcolor{LightCyan}July	&	18	&	47	&	forum264LIWC	&	0.38	&	1.00	&	0.55	\\
\rowcolor{LightCyan}July	&	18	&	50	&	forum250LIWC	&	0.36	&	1.00	&	0.53	\\
July	&	18	&	43	&	baseline	&	0.37	&	0.89	&	0.52	\\
\rowcolor{LightCyan}July	&	18	&	35	&	forum8senti	&	0.37	&	0.72	&	0.49	\\
\rowcolor{LightCyan}July	&	18	&	50	&	forum111LIWC	&	0.32	&	0.89	&	0.47	\\
\hline
August	&	28	&	39	&	baseline	&	0.67	&	0.93	&	0.78	\\
\rowcolor{LightCyan}August	&	28	&	31	&	forum264LIWC	&	0.65	&	0.71	&	0.68	\\
\rowcolor{LightCyan}August	&	28	&	32	&	forum121LIWC	&	0.63	&	0.71	&	0.67	\\
\rowcolor{LightCyan}August	&	28	&	35	&	forum211vader	&	0.60	&	0.75	&	0.67	\\
\rowcolor{LightCyan}August	&	28	&	33	&	forum194LIWC	&	0.61	&	0.71	&	0.66	\\
\hline
September	&	31	&	40	&	baseline	&	0.60	&	0.77	&	0.68	\\
\rowcolor{LightCyan}September	&	31	&	38	&	forum210senti	&	0.61	&	0.74	&	0.67	\\
\rowcolor{LightCyan}September	&	31	&	37	&	forum121LIWC	&	0.57	&	0.68	&	0.62	\\
\rowcolor{LightCyan}September	&	31	&	46	&	forum219vader	&	0.50	&	0.74	&	0.60	\\
\rowcolor{LightCyan}September	&	31	&	30	&	forum194LIWC	&	0.60	&	0.58	&	0.59	\\
\hline
\rowcolor{LightCyan}October	&	53	&	44	&	forum210LIWC	&	0.77	&	0.64	&	0.70	\\
October	&	53	&	47	&	baseline	&	0.74	&	0.66	&	0.70	\\
\rowcolor{LightCyan}October	&	53	&	41	&	forum264LIWC	&	0.78	&	0.60	&	0.68	\\
\rowcolor{LightCyan}October	&	53	&	39	&	forum250LIWC	&	0.74	&	0.55	&	0.63	\\
\rowcolor{LightCyan}October	&	53	&	40	&	forum8vader	&	0.73	&	0.55	&	0.62	\\
\hline
November	&	37	&	52	&	daywise-baserate	&	0.62	&	0.86	&	0.72	\\
\rowcolor{LightCyan}November	&	37	&	49	&	forum121LIWC	&	0.57	&	0.76	&	0.65	\\
\rowcolor{LightCyan}November	&	37	&	53	&	forum147LIWC	&	0.55	&	0.78	&	0.64	\\
\rowcolor{LightCyan}November	&	37	&	50	&	forum111LIWC	&	0.56	&	0.76	&	0.64	\\
\rowcolor{LightCyan}November	&	37	&	50	&	forum194LIWC	&	0.56	&	0.76	&	0.64	\\
\hline
December	&	35	&	30	&	daywise-baserate	&	0.67	&	0.57	&	0.62	\\
December	&	35	&	27	&	baseline	&	0.63	&	0.49	&	0.55	\\
\rowcolor{LightCyan}December	&	35	&	23	&	forum250LIWC	&	0.65	&	0.43	&	0.52	\\
\rowcolor{LightCyan}December	&	35	&	28	&	forum194LIWC	&	0.57	&	0.46	&	0.51	\\
\rowcolor{LightCyan}December	&	35	&	29	&	forum147LIWC	&	0.55	&	0.46	&	0.50	\\
\hline
January	&	43	&	42	&	baseline	&	0.60	&	0.58	&	0.59	\\
January	&	43	&	37	&	daywise-baserate	&	0.59	&	0.51	&	0.55	\\
\rowcolor{LightCyan}January	&	43	&	35	&	forum219vader	&	0.60	&	0.49	&	0.54	\\
\rowcolor{LightCyan}January	&	43	&	37	&	forum111LIWC	&	0.57	&	0.49	&	0.53	\\
\rowcolor{LightCyan}January	&	43	&	37	&	forum147LIWC	&	0.57	&	0.49	&	0.53	\\

\hline%

\end{tabular}%
\end{table}

\begin{table}[!t]
\centering
\caption{Results from Organization B's Malicious-Destination
\label{tab:output5}} 
%\begin{tabular}{llccccc}%
\begin{tabular}{llccccc}
\hline%
\textbf{Month}&\textbf{Evt}&\textbf{Warn}&\textbf{Signal}&\textbf{P}&\textbf{R}&\textbf{F1}\\%
\hline%

\rowcolor{LightCyan}July	&	6	&	8	&	forum130vader	&	0.63	&	0.83	&	0.71	\\
\rowcolor{LightCyan}July	&	6	&	8	&	forum8senti	&	0.63	&	0.83	&	0.71	\\
\rowcolor{LightCyan}July	&	6	&	8	&	forum111LIWC	&	0.50	&	0.67	&	0.57	\\
\rowcolor{LightCyan}July	&	6	&	12	&	forum194LIWC	&	0.42	&	0.83	&	0.56	\\
\rowcolor{LightCyan}July	&	6	&	9	&	forum210senti	&	0.44	&	0.67	&	0.53	\\
\hline
\rowcolor{LightCyan}August	&	8	&	6	&	forum210senti	&	0.67	&	0.50	&	0.57	\\
August	&	8	&	17	&	daywise-baserate	&	0.35	&	0.75	&	0.48	\\
\rowcolor{LightCyan}August	&	8	&	13	&	forum211senti	&	0.38	&	0.63	&	0.48	\\
\rowcolor{LightCyan}August	&	8	&	5	&	forum210LIWC	&	0.60	&	0.38	&	0.46	\\
\rowcolor{LightCyan}August	&	8	&	21	&	forum8vader	&	0.29	&	0.75	&	0.41	\\
\hline
September	&	6	&	11	&	daywise-baserate	&	0.55	&	1.00	&	0.71	\\
\rowcolor{LightCyan}September	&	6	&	9	&	forum210LIWC	&	0.56	&	0.83	&	0.67	\\
\rowcolor{LightCyan}September	&	6	&	10	&	forum250LIWC	&	0.30	&	0.50	&	0.37	\\
\rowcolor{LightCyan}September	&	6	&	11	&	forum121LIWC	&	0.27	&	0.50	&	0.35	\\
\rowcolor{LightCyan}September	&	6	&	1	&	forum147LIWC	&	1.00	&	0.17	&	0.29	\\
\hline
October	&	9	&	8	&	daywise-baserate	&	0.25	&	0.22	&	0.24	\\
\rowcolor{LightCyan}October	&	9	&	2	&	forum121LIWC	&	0.50	&	0.11	&	0.18	\\
\rowcolor{LightCyan}October	&	9	&	114	&	forum210senti	&	0.03	&	0.33	&	0.05	\\
October	&	9	&	0	&	arimax	&	0.00	&	0.00	&	0.00	\\
October	&	9	&	0	&	dark-mentions	&	0.00	&	0.00	&	0.00	\\
\hline
November	&	4	&	14	&	daywise-baserate	&	0.29	&	1.00	&	0.44	\\
\rowcolor{LightCyan}November	&	4	&	5	&	forum210LIWC	&	0.20	&	0.25	&	0.22	\\
\rowcolor{LightCyan}November	&	4	&	21	&	forum219vader	&	0.10	&	0.50	&	0.16	\\
\rowcolor{LightCyan}November	&	4	&	9	&	forum211vader	&	0.11	&	0.25	&	0.15	\\
\rowcolor{LightCyan}November	&	4	&	13	&	forum210senti	&	0.08	&	0.25	&	0.12	\\
\hline
December	&	3	&	12	&	daywise-baserate	&	0.17	&	0.67	&	0.27	\\
December	&	3	&	0	&	arimax	&	0.00	&	0.00	&	0.00	\\
December	&	3	&	0	&	dark-mentions	&	0.00	&	0.00	&	0.00	\\
December	&	3	&	0	&	deep-exploit	&	0.00	&	0.00	&	0.00	\\
\hline
January	&	5	&	18	&	daywise-baserate	&	0.22	&	0.80	&	0.35	\\
January	&	5	&	0	&	arimax	&	0.00	&	0.00	&	0.00	\\
January	&	5	&	0	&	dark-mentions	&	0.00	&	0.00	&	0.00	\\
January	&	5	&	0	&	deep-exploit	&	0.00	&	0.00	&	0.00	\\

\hline%

\end{tabular}%
\end{table}

\begin{table}[!t]
\centering
\caption{Results from Organization B's Malicious-Email
\label{tab:output6}} 
\begin{tabular}{llccccc}
\hline%
\textbf{Month}&\textbf{Evt}&\textbf{Warn}&\textbf{Signal}&\textbf{P}&\textbf{R}&\textbf{F1}\\%
\hline%

\rowcolor{LightCyan}July	&	24	&	49	&	forum210LIWC	&	0.33	&	0.67	&	0.44	\\
\rowcolor{LightCyan}July	&	24	&	56	&	forum210senti	&	0.30	&	0.71	&	0.43	\\
July	&	24	&	75	&	baseline	&	0.23	&	0.71	&	0.34	\\
July	&	24	&	81	&	daywise-baserate	&	0.21	&	0.71	&	0.32	\\
\rowcolor{LightCyan}July	&	24	&	81	&	forum130vader	&	0.21	&	0.71	&	0.32	\\
\hline
\rowcolor{LightCyan}August	&	57	&	55	&	forum111LIWC	&	0.55	&	0.53	&	0.54	\\
August	&	57	&	70	&	baseline	&	0.49	&	0.60	&	0.54	\\
August	&	57	&	91	&	daywise-baserate	&	0.43	&	0.68	&	0.53	\\
\rowcolor{LightCyan}August	&	57	&	107	&	forum147LIWC	&	0.39	&	0.74	&	0.51	\\
\rowcolor{LightCyan}August	&	57	&	153	&	forum6senti	&	0.33	&	0.88	&	0.48	\\
\hline
September	&	179	&	70	&	daywise-baserate	&	0.76	&	0.30	&	0.43	\\
\rowcolor{LightCyan}September	&	179	&	102	&	forum210senti	&	0.58	&	0.33	&	0.42	\\
\rowcolor{LightCyan}September	&	179	&	180	&	forum210LIWC	&	0.40	&	0.40	&	0.40	\\
\rowcolor{LightCyan}September	&	179	&	100	&	forum147LIWC	&	0.54	&	0.30	&	0.39	\\
September	&	179	&	76	&	baseline	&	0.57	&	0.24	&	0.34	\\
\hline
October	&	71	&	125	&	daywise-baserate	&	0.50	&	0.87	&	0.63	\\
October	&	71	&	118	&	baseline	&	0.49	&	0.82	&	0.61	\\
\rowcolor{LightCyan}October	&	71	&	90	&	forum211senti	&	0.53	&	0.68	&	0.60	\\
\rowcolor{LightCyan}October	&	71	&	142	&	forum194LIWC	&	0.44	&	0.89	&	0.59	\\
\rowcolor{LightCyan}October	&	71	&	150	&	forum210senti	&	0.42	&	0.89	&	0.57	\\
\hline
November	&	426	&	104	&	daywise-baserate	&	0.67	&	0.16	&	0.26	\\
\rowcolor{LightCyan}November	&	426	&	205	&	forum264LIWC	&	0.39	&	0.19	&	0.25	\\
November	&	426	&	118	&	baseline	&	0.55	&	0.15	&	0.24	\\
\rowcolor{LightCyan}November	&	426	&	251	&	forum210LIWC	&	0.31	&	0.18	&	0.23	\\
\rowcolor{LightCyan}November	&	426	&	579	&	forum210senti	&	0.20	&	0.27	&	0.23	\\
\hline
\rowcolor{LightCyan}December	&	51	&	69	&	forum210LIWC	&	0.30	&	0.41	&	0.35	\\
\rowcolor{LightCyan}December	&	51	&	329	&	forum147LIWC	&	0.09	&	0.55	&	0.15	\\
\rowcolor{LightCyan}December	&	51	&	313	&	forum111LIWC	&	0.08	&	0.51	&	0.14	\\
\rowcolor{LightCyan}December	&	51	&	249	&	forum194LIWC	&	0.08	&	0.41	&	0.14	\\
\rowcolor{LightCyan}December	&	51	&	284	&	forum211senti	&	0.08	&	0.45	&	0.14	\\
\hline
January	&	10	&	12	&	deep-exploit	&	0.25	&	0.30	&	0.27	\\
January	&	10	&	103	&	daywise-baserate	&	0.10	&	1.00	&	0.18	\\
January	&	10	&	186	&	baseline	&	0.05	&	1.00	&	0.10	\\
\rowcolor{LightCyan}January	&	10	&	226	&	forum111LIWC	&	0.04	&	1.00	&	0.08	\\
% January	&	10	&	228	&	forum121LIWC	&	0.04	&	1.00	&	0.08	\\

\hline%

\end{tabular}%
\end{table}

%% file: src/rw.tex
\section{Related Work}

Given the serious nature of cyber attacks, naturally there are a number of other research efforts to predict such attacks. As it relates to our efforts, the three main areas of research are sentiment analysis in cyber security, predictive methods for cyber attacks and leveraging dark web data in cyber security.

\subsection{Sentiment Analysis in Cyber Security}

The closest work which has applied sentiment analysis of hacker forums to cyber security is \cite{macdonald2015identifying}. While much research has investigated the specifics of cyber attacks, \cite{macdonald2015identifying} investigates the actual cyber actors via their communication activities. Their focus was the cyber-physical systems related to critical infrastructure and they developed an automated analysis tool to identify potential threats against such infrastructure. Despite recognizing that there are over 140 hacker forums on the public web, the authors chose only one forum to scrape the complete forum once. They leveraged the Open Discussion Forum Crawler to do the scrapping and then used OpenNLP to tag parts of speech, filtering on nouns. Those nouns were cross referenced with three list of malicious keywords to identify posts whose sentiment would be determined with SentiStrength. Contextual analysis of keyword pairings with sentiment scores allowed them to confirm current statistics about critical infrastructure cyber attacks. The main differences illustrated in our work is that we use looked at over 100 forums, not just one from both the Dark Web and Surface Web. In collecting posts for over a two year period, we found the sentiment of all posts applying Vader and LIWC for sentiment in addition to just SentiStrength. Furthermore, we were able to model our data against ground truth events from companies making our approach predictive in nature. 

BiSAL \cite{al2015bisal} did sentiment analysis in English and Arabic on Dark Web forums with slight modification to cyber security terms. Other work such as \cite{chen2008sentiment} use sentiment in measuring radicalization. Remaining research in sentiment analysis, not specific to cyber security was presented earlier.

\subsection{Predicting Cyber Attack}

The issue of predicting cyber attacks is not new and their has been a considerable research effort in this field. The efforts split along two categories, using network traffic or non-network traffic. Forecasting methods such as \cite{park2012cyber,pontes2011applying,leslie2017statistical} analyze network traffic. Where \cite{park2012cyber} is specific to predicting attacks using IPV4 packet traffic, and  \cite{pontes2011applying} looks at various network sensors at different layers to prevent unwanted Internet traffic, whereas \cite{leslie2017statistical} combines DNS traffic with security metadata such as number of policy violations and the number of clients in the network. Many researchers such as \cite{zhang2015predicting} based cyber prediction on open source information. They use the National Vulnerability Database. They highlight the difficulty in using public sources for building effective models. Other work has focused on detecting cyber bullying using graph detection models \cite{nahar2012sentiment} with success, but is limited in malicious activity and not a predictive model. 

The closest to our research is Gandotra et al \cite{comptech} who outlined a number cyber prediction efforts using statistical modeling and algorithmic modeling. They highlight several significant challenges that we tried to address. The first challenge is that open source ground truth is incomplete and should be compiled from multiple sources and analysis doesn't scale to real world scenarios.  We were able to get ground truth data from 2 companies that operate in the defense industrial base, this ground truth is across three different attack vectors and is over a two year time period. The additional challenges in \cite{comptech} focus on the volume, speed and heterogeneity of network data which we avoid since we are attempting to prevent cyber events specifically with non-network data. They also present two modeling approaches of statistical modeling and algorithmic modeling. We used statistical models not unlike what they present as classical time series models with auto-regressive, integrated moving average with historical data and external signals.

\subsection{Dark Web Research}

There has been a lot of research recently concerning the Dark Web or websites not indexed by major search engines. Typically the Dark Web refers to the TOR \cite{dingledine2004tor} network which is only accessible via specialized browsers. It has been shown by \cite{nunes2016darknet} that from an overall cyber security threat perspective, the Dark Web provides a valuable source of information for malicious activity. They developed a system that scrapes hacker forum and marketplace sites on the Dark Web to develop threat warnings for cyber defenders. We leverage the same data source but perform sentiment analysis to not only predict future threats, but to predict actual attacks. They also leverage the Deep Net which is the portion of the Surface Web not indexed by standard search engines.   

While not using sentiment analysis, \cite{lacey2015s} offers insight to the trust establishment between participants in Dark Web forums. There may be behavioral patterns of malicious actors that provide insight to future activity. Dark Web conversations were shown to provide earlier insights than Surface Web conversations by \cite{sapienza2017early}  indicating potential predictive power for cyber events. \cite{sapienza2017early} highlight two cases with a major DDoS attack and the Mirai attack. There may also be early insights on the Surface Web in many of the social media sites as illustrated in \cite{sabottke2015vulnerability}. Our work focused only on forums where it was likely that computer security items would be discussed, but does contain a mix of Dark Web and Surface Web. There has been work using natural language processing on Dark Web text for predictive method such as \cite{tavabi2018dark}. Other predictive approaches such as Cyber Attacker Model Profile (CAMP) \cite{watters2012characterising}, focus on the macro level of a country and financial cyber crimes, where we look at a wider range of malicious activity against specific target organizations.

%% file: src/concl.tex
\section{Conclusions}
Malicious activity can be very devastating to national security, economies, businesses and personal lives. As such, cyber security professionals working with major organizations and nation states could use all the help they can get in preventing malicious activity. We present a methodology to predict malicious cyber events by exploiting malicious actor's behavior via sentiment analysis of posts on hacker forums. These forums on both Surface Web and Dark Web have some predictive power to be used as signals external to the network for forecasting attacks using time series models. Using ground truth data from two major organizations in the Defense Industrial Base across three different cyber event types, we show that sentiment signals can be more predictive than a baseline time series model. Additionally, they will often beat other state of the art external signals, in the 7 months under study across the 3 event types from the 2 organizations, sentiment signals performed the best 15 out of 42 times or 36\%.  The signal parameters need to be tuned over significant historical data and the source forum could be shut off or taken down at any time; however, an automated implementation of this system would still be value added. 